# Review

# Evolving the Behavior of Machines: From Micro to Macroevolution

Jean-Baptiste Mouret[1,*]

**SUMMARY**

**Evolution gave rise to creatures that are arguably more sophisticated than the greatest human-designed systems. This feat has inspired computer scientists since the advent of computing and led to optimization tools that can evolve complex neural networks for machines—an approach known as "neuroevolution." After a few successes in designing evolvable representations for high-dimensional artifacts, the field has been recently revitalized by going beyond optimization: to many, the wonder of evolution is less in the perfect optimization of each species than in the creativity of such a simple iterative process, that is, in the diversity of species. This modern view of artificial evolution is moving the field away from microevolution, following a fitness gradient in a niche, to macroevolution, filling many niches with highly different species. It already opened promising applications, like evolving gait repertoires, video game levels for different tastes, and diverse designs for aerodynamic bikes.**

## INTRODUCTION

Evolution by natural selection is the master algorithm of life: an infinite variation/selection loop that gave rise to the astonishing diversity of life-forms that inhabit our planet. That such an apparently simple iterative process is at the origin of so much sophistication has fascinated computer scientists since the advent of computers. Starting from the 1960s, several groups took inspiration from evolutionary biology to develop "artificial evolution" algorithms. They converged to modern "evolutionary algorithms" (De Jong, 2016). Given a representation for possible solutions (a list of numbers [De Jong, 2016], a graph [Sims, 1994], a neural network [Stanley and Miikkulainen, 2002], a program [Koza, 1992], and so forth) and a *fitness function* that measures their performance at the task, all variants loop over the same three steps:

(1) evaluate the fitness of each individual of the population (*evaluation*);

(2) rank then select the individual using their fitness value (*selection*);

(3) apply variation operators on the best individuals to create a new population (*variation*).

The process is bootstrapped by generating an initial population randomly. Depending on the variant, a new population is created at each iteration (non-elitist algorithms) or offspring compete with their parents to stay in the population (elitist algorithms). Two variation operators are used: mutation and crossover. Mutation consists in adding random variations to a single genome; for instance, if the genome is a list of real numbers, mutation can be implemented by adding Gaussian noise to these numbers (in current algorithms, self-adjusting perturbations are used [Hansen et al., 2003]). Crossover consists in mixing two genomes of the population, in the hope of combining their features; in the case of a list of numbers, this can be implemented by adding a linear combination of the elements of the "parents" (Deb and Beyer, 2001) (depending on the representation, crossover is not always used).

From the perspective of computer science, artificial evolution is currently considered as a mathematical optimization algorithm, that is, as an algorithm that finds the maximum of a function. Such algorithms have countless applications in engineering, machine learning, bioinformatics, logistics, etc. (Kochenderfer and Wheeler, 2019) because many problems can be formalized as the maximization (or minimization) of a numerical objective. In the vast landscape of optimization algorithms, evolutionary algorithms are a good

[1]Inria, CNRS, Université de Lorraine, LORIA, Nancy 54000, France

*Correspondence: jean-baptiste.mouret@inria.fr

https://doi.org/10.1016/j.isci.2020.101731

approach for functions for which the analytical gradient cannot be computed, called *black-box functions*. In particular, variants of the Covariance Matrix Adaptation Evolutionary Strategy (CMA-ES) (Hansen et al., 2003) currently outperform other black-box optimization algorithms. Evolutionary algorithms are also most suited to parallel processing computers, like modern multicore machines and computing clusters, because each candidate solution of the population can be evaluated in parallel.

From the perspective of biology, evolutionary algorithms can be viewed as a model of natural evolution: all individuals compete against each other, the fittest reproduce and least fit disappear; but repeated application of this simple rule of natural selection leads these algorithms to converge on a single genome, that of the fittest individual, rather than the diversity of species that we observe in nature. Such convergence implies that traditional evolutionary algorithms more closely resemble a microevolutionary process (Reznick and Ricklefs, 2009; Erwin, 2000), that is, evolution *within* a single ecological niche in which all individuals compete together, which most of the time results in the dominance of a single species. By contrast, macroevolution is the appearance of new species that occupy different niches and explains the diversity of species as a consequence of the diversity of niches (Reznick and Ricklefs, 2009; Erwin, 2000). This article will show that evolutionary algorithms that more closely follow the principles of macroevolution have greater benefit in the evolution of the behavior of machines.

At first sight, mathematical optimization and microevolution might seem far from the behavior of machines. Instead, searching for a ''brain'' of a machine is often viewed as a reinforcement learning problem (Sutton and Barto, 2018), in which the objective is to learn a *policy* that allows the agent to maximize its expected *reward* by exploring its environment. This kind of algorithm recently gained much interest in the artificial intelligence community thanks to impressive results in *deep reinforcement learning*, like learning to play two-dimensional video games from pixels (Mnih et al., 2015).

Nevertheless, evolutionary algorithms are a compelling alternative to discover policies for intelligent machines (Stanley et al., 2019; Whiteson, 2012). To evolve a policy, the fitness function is defined as the sum of rewards during an episode: each policy is run for some time on the system (simulated or real) and the rewards are accumulated. Concerning the policy itself, most of the efforts have been focused on evolving artificial neural networks because they are theoretically capable of representing any function and because they have inspiring connections to biology (Floreano et al., 2008; Pfeifer and Bongard, 2006; Stanley et al., 2019). These neural networks take sensor readings as inputs and compute the motor values (wheel velocities, joint positions, etc.) as outputs. Importantly, they usually do not learn during their lifetime, as the weights are fixed at the beginning of each evaluation of the fitness, although a few studies investigated the evolution of plastic neural networks (Floreano et al., 2008; Mouret and Tonelli, 2014; Soltoggio et al., 2018). This means that ''learning'' usually happens at the phylogenetic timescale, from generation to generation (Togelius et al., 2009).

Compared with classic reinforcement learning, this *neuroevolution* approach considers the value of the policy without attempting to identify the most useful state-action pairs. Evolutionary methods therefore discard potentially useful information contained in the sequence of decisions and states followed by the agent. However, by doing so, they bypass many of the challenges of reinforcement learning (Whiteson, 2012; Togelius et al., 2009). For instance, they find policies when the state space is large or continuous (or both), especially where there exist simple reactive strategies like those followed by insects (Braitenberg, 1986; Nolfi et al., 2000; Pfeifer and Bongard, 2006). In addition, neuroevolution handles naturally the *credit assignment problem* (Sutton and Barto, 2018) since it does not attempt to identify which decision leads to a high reward.

Neuroevolution raises its own challenges, however. The present article is focused on the two most daunting ones: the *representation problem*—''How can an artificial genotype describe a neural network that could be as complex as a brain?''—and the *stepping stones problem*—''what intermediate steps lead to an artifact as sophisticated as a brain?''

## NEUROEVOLUTION: EVOLVING ARTIFICIAL BRAINS

The most direct way to evolve a neural network is to fix the topology, typically either a feedforward multi-layer perceptron or a recurrent neural network, and only evolve the weights (Dasgupta and McGregor, 1992; Nolfi et al., 2000). In this case, the genotype is a large list of floating-point numbers. Mutation can be a simple Gaussian perturbation applied to randomly chosen weights. The crossover operator is more difficult to define because two neural networks can be topologically identical but ordered differently, which results in meaningless crossovers. In addition, neural networks typically distribute their "knowledge" in all the weights, which means that crossing two parents is unlikely to produce a neural network that combine their functionality. As a consequence, most authors choose to not use the crossover operators and only use random mutations.

Despite its simplicity, this approach gives competitive results in many benchmarks. For instance, the CMA-ES algorithm (Hansen et al., 2003) can evolve the weights of small neural networks (for example, four inputs/ two hidden neurons/one output) for simple control tasks, like balancing a pole on a cart, with an order of magnitude fewer episodes than many of the classic reinforcement learning algorithms (Heidrich-Meisner and Igel, 2009). Perhaps more surprisingly, simple evolutionary algorithms can also evolve the weights of deep neural networks with millions of weights (Such et al., 2017; Salimans et al., 2017; Mania et al., 2018). In these experiments, the authors consider the same benchmark tasks as those used to validate state-of-the-art deep reinforcement learning algorithms, in particular playing Atari games from pixels (Mnih et al., 2015) and continuous control like learning to walk with a simulated manikin (Heess et al., 2017). They also exploit the same deep convolutional neural networks that proved successful in deep reinforcement learning. The results show that neuroevolution can reach similar scores as state-of-the-art deep reinforcement learning algorithms. In addition, although neuroevolution requires more computation, it runs faster (in term of wall-clock time) when many computer cores are available because it is straightforward to compute the fitness values in parallel (Such et al., 2017; Salimans et al., 2017).

The perfect topology for a neural network is, however, rarely known. The problem might, for example, require recurrent connections to take time into account, or it might need more hidden neurons, or there might exist solutions with order of magnitudes fewer connections. A natural extension of evolving the weights is therefore to create variation operators that add/remove neurons and connections; as there is a direct and bidirectional link between the genotype (the lists or matrix of connections, neurons, and weights) and the phenotype (the neural network), this is called a *direct encoding*. Unfortunately, adding new neurons or connections is very disruptive for the functionality of the network, which makes the new networks lose the competition with the previous, better tuned structures, which in turn, prevents any structural innovation. The influential Neuroevolution of Augmented Topologies (NEAT) approach (Stanley and Miikkulainen, 2002) addresses this issue by dividing the population into "species" with similar topologies, so that only similar networks are in competition. In practice, NEAT compares networks by using a historical tracking mechanism that makes it easy to identify the common parts of two networks from the population. Among other successes, NEAT has been successful in finding "minimal" neural networks for control tasks (Stanley and Miikkulainen, 2002), to train non-player characters in video games (Stanley et al., 2005), and it even helped in physics to discover the most accurate measurement yet of the mass of the top quark (Aaltonen et al., 2009).

Direct encodings like NEAT are effective for simple tasks but are unlikely to eventually scale up to neural networks as large as a brain (billions of neurons) because they would need to keep rediscovering interesting structures each time they would create a new connection. By contrast, natural organisms reuse many genes, which makes it possible to apply the same solution many times. For example, the same genes are used for both eyes, for every finger of the hand, etc.: only about 25,000 genes are enough to describe the trillions of cells that make up a human (Southan, 2004). This feat is possible because the relationship between genes and biological networks is not one-to-one; instead, genes are *developed* to the phenotype through a complex *genotype to phenotype map* (Ahnert, 2017), which is still little understood. As a result of this developmental process, natural networks tend to be *modular* and *regular* (Wagner et al., 2007; Lipson, 2007; Clune et al., 2013), which means that they separate functions into structural units (modules) that can be reused, with or without variation (for example, symmetries are repetition with variations). This line of thought encouraged the evolutionary computation community to propose *developmental encodings* (also called *indirect encodings*), which are compressed representations of networks that make it possible to create modular and regular phenotypes from shorter descriptions (Stanley and Miikkulainen, 2003).

The first developmental encodings were inspired by grammatical models of the biological development of multi-cellular organisms (Lindenmayer, 1968). In these systems, a formal grammar defines a set of rewriting rules that are applied recursively on a starting pattern. For instance, taking the rules $A \rightarrow AB$, $B \rightarrow BB$ and starting with *AB*, the genome would develop to *ABBB*, then, recursively, to *ABBBBBBB*, etc. (until a maximum depth is reached). This set of rules can be evolved, which means that the genome is the grammar, and the resulting string can be interpreted as instructions to build a structure with a LOGO-like language (A = forward, B = turn left, C = divide in two, …) (Kitano, 1990; Gruau, 1994; Kodjabachian and Meyer, 1998; Hornby and Pollack, 2001). The most iconic work in this direction is certainly Karl Sims' work evolving both the morphology and the controller of artificial creatures embedded in a physically realistic simulation (Sims, 1994), whose videos influenced generations of scientists in artificial evolution. A variant of this idea is to evolve a development program that follows a predefined grammar (Mouret and Doncieux, 2008; Miller, 2003), which connects developmental encodings to genetic programming (Banzhaf et al., 1998) (evolving computer programs or algorithms).

With the objective of being closer to biology, the cell chemistry approaches takes inspiration from reaction-diffusion models (Turing, 1952) and gene regulation networks (Davidson, 2010; Cussat-Blanc et al., 2019). In these models, genes produce proteins that either have a direct phenotypic effect or regulate the expression of other genes (Bongard and Pfeifer, 2003; Cussat-Blanc et al., 2019); in some implementations, the proteins can be embedded in a Cartesian space (Cangelosi et al., 1994; Eggenberger, 1997) to take physical proximity into account. Overall, these approaches can give useful insights for evolutionary biology and have attractive properties, like the ability for an evolved digital organism to "self-repair" (Cussat-Blanc et al., 2019). However, they typically require arbitrarily choosing many parameters, which makes them challenging to use in practice.

A perfect indirect encoding needs to be able to express the design motifs that we observe in nature, like symmetries (for example, left-right symmetry in most animals), symmetries with variation (for example, fingers), and repetition (for example, skin cells), while being abstract enough to not be encumbered by unnecessary parameters or hypotheses. Taking inspiration from evolutionary art (Sims, 1991) and functional composition in genetic programming (Koza, 1992), Compositional Pattern Producing Networks (CPPN) attempts to achieve this balance by describing the spatial relationships that result from the developmental process without simulating the process itself (Stanley, 2007). The general idea is to model development as a function of coordinates in a spatial substrate, that is, $p_{xy} = f(x,y)$, where $p$ is the phenotype (for example, a gray level) at coordinate $x$ and $y$. To obtain the phenotype, the function $f$ is called for every coordinate of the substrate. The properties of $f$ give rises to many interesting patterns. For example, if $f(x,y)$ is symmetric along the x axis, that is, if $f(x,y) = f(-x,y)$, then the phenotype will be symmetric along the x axis; similarly, if $f$ is periodic, for instance, $f(x,y) = sin(x \times y)$, then repetitive patterns will emerge; more generally, $f$ can be a composition of functions, like $f(x,y) = sin(abs(x^3))$, that can encode arbitrarily complex spatial relationships.

These compositional networks can be described as a network of functions, which is equivalent to a feedforward neural networks for which each neuron can have different activation function (for example, $sinx(x)$, $abs(x)$). As a consequence, compositional networks can be evolved with existing direct encodings like NEAT (Stanley, 2007). To explore the potential of this representation, an interactive evolution website was setup (http://www.picbreeder.org), which shows many striking examples of life-like shapes (Secretan et al., 2011).

Besides images, Compositional pattern producing networks can describe the connectivity patterns of neural networks (Stanley et al., 2009; Clune et al., 2011) if they are queried for a four-dimensional space: to compute the weight between two neurons that are both in a two-dimensional space, a network with four inputs is used (for example, $w_{1,2} = f(x_1, y_1, x_2, y_2)$). This indirect encoding is called HyperNEAT, because it evolves a network with the direct encoding NEAT (Stanley and Miikkulainen, 2002) to describe patterns in a four-dimensional hypercube. As the same network $f$ can be queried for million of neurons, HyperNEAT can evolve neural networks with millions of connections while compressing the information in a much simpler network, for instance, to play Atari games (Hausknecht et al., 2014), make simulated creatures walk (Clune et al., 2011), or play checkers (Gauci and Stanley, 2008).

Most of the direct and indirect encodings have been extended to describe not only neural controllers but also the mechanical design of robots. In that case, like in nature, the brain and the body co-evolve so that they complement each other (Pfeifer and Bongard, 2006). Examples include evolving structures made of bars and linear actuators (Lipson and Pollack, 2000), assemblies of cubic modules with and without actuators (Jelisavcic et al., 2017; Brodbeck et al., 2015), and more recently, living cells (Kriegman et al., 2020). For

> **Box 1. The Reality Gap Problem**
>
> Evolution on Earth took millions of years while concurrently evaluating the fitness of billions of individuals. Such a scale is hard to envision with real machines, but it could be simulated on computers (at least to some extent). This is why most work in artificial evolution is performed in simulation, with the hope of ultimately transferring the result to a real system.
>
> Unfortunately, results found in simulation rarely work well on the real system, mainly because evolution (like reinforcement learning) blindly exploits all the inaccuracies of the simulation. This leads to thousand of examples of "surprising" solutions (Lehman et al., 2020) that could not happen in the real world. This is the "Reality Gap" problem (Jakobi, 1997), which is nowadays called "Sim2Real" in the reinforcement learning literature.
>
> Over the years, many approaches have been proposed to circumvent this obstacle, which can often be combined:
>
> - evolve directly on real robots and evaluate the fitness with motion capture systems (Nolfi et al., 2000; Hornby et al., 2005; Brodbeck et al., 2015);
> - learn a surrogate model (a neural network or Gaussian processes) that predicts the fitness values and use it in the algorithm instead of the fitness function, which is sometimes called Bayesian optimization (Gaier et al., 2018; Shahriari et al., 2015; Cully et al., 2015);
> - improve the simulator either with extensive tuning (Tan et al., 2018) or by using data to improve it automatically (Bongard et al., 2006), which is called "system identification" in robotics;
> - Make the policies (controllers) more robust to different dynamics by:
>   - adding noise to the sensors and actuators (Jakobi, 1997), which is nowadays called "Domain Randomization" (Tobin et al., 2017);
>   - letting controllers learn during their lifetime by evolving plastic neural networks (Floreano and Mondada, 1998; Floreano et al., 2008; Soltoggio et al., 2018);
> - Take the limits of the simulation into account by learning to predict which controllers will transfer well (Koos et al., 2012);
> - Evolve many different solutions with a quality diversity (Cully et al., 2015) or multi-objective algorithm (Penco et al., 2020), then test on the real system to find those that work best on the real robot.

practical reasons, evolution is typically performed in simulation, and only the final design is manufactured. Nevertheless, transfer from the simulation to the real world is always challenging, even when only neural controllers are evolved (Box 1).

## THE STEPPING STONES PROBLEM AND NOVELTY SEARCH

When we use indirect encodings, we implicitly assume that the evolutionary process will exploit the interesting possibilities offered by the encoding. For example, if the representation can define and reuse modules, we expect the evolutionary process to discover, use, and reuse modules. Unfortunately, this is rarely observed in experiments, although this often goes unreported. The main reason is a central question in evolutionary biology: evolution selects according to short-term advantages, whereas many traits, like the modularity of the phenotype, only provide a selective advantage on the long term (Gould and Vrba, 1982; Pigliucci, 2008). As a consequence, modularity, repetition, and hierarchy need to be by-products of other selective pressures, like the connection cost (Clune et al., 2013) or the necessity to react to quickly changing environments (Kashtan and Alon, 2005). Interestingly, this modularity does not need any special genotype and can appear both with direct (Clune et al., 2013) and indirect encodings (Huizinga et al., 2014).

In hindsight, the manipulation of selective pressures in evolutionary algorithms has not been explicit for a long time (Doncieux and Mouret, 2014). This might be because of the optimization-centric view that dominated the field, according to which the fitness function is an untouchable, user-defined input to the evolutionary algorithm (Doncieux and Mouret, 2014). Or maybe because the representation seemed to be a more fundamental problem. At any rate, it is remarkable to observe a recent shift in interest from encodings to selective pressures (Doncieux and Mouret, 2014), although early successes like the NEAT encoding (Stanley and Miikkulainen, 2002) are actually more about controlling the selective pressure (to protect novel structures) than the encoding itself.

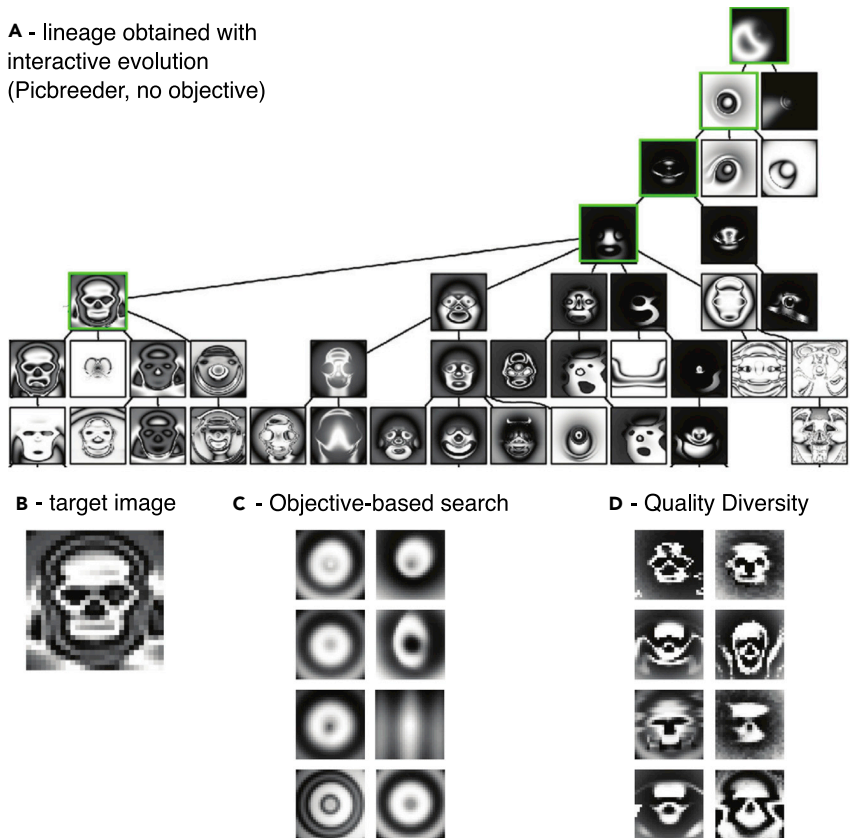

**Figure 1. The Stepping Stones Problem**

(A) Example of a "family tree" in Picbreeder, evolved with interactive evolution without any explicit objective (Secretan et al., 2011). The stepping stones to reach a specific image (for example, the skull on the left) do not look like like the final image. This explains why objective-based search is unable to find the same picture by setting a specific picture as a target (Woolley and Stanley, 2011).
(B) Image from Picbreeder set as a target image for objective-based search.
(C) Typical results found with objective-based search (Woolley and Stanley, 2011; Gaier et al., 2019).
(D) Typical results found with MAP-Elites (Mouret and Clune, 2015), a Quality Diversity algorithm (see the next section), adapted from (Gaier et al., 2019).

The challenge was made explicit when analyzing the result of the Picbreeder interactive evolutionary experiment, which we described before (Secretan et al., 2011). In this experiment, Internet users were presented with a set of images encoded with Compositional Pattern Producing Networks (Stanley, 2007), asked to click on the ones that they like best, and the system called genetic operators to generate new images. After a few generations by hundreds of users, the results show a large range of interesting images (Figure 1A), which demonstrates the potential of the encoding. A follow-up study attempted to evolve an image found with Picbreeder without any human input (Woolley and Stanley, 2011). In that case, a standard evolutionary algorithm was set to find an image as close as possible to the target, using the same encoding and the same variation operators. Surprisingly, nothing close was ever found (Woolley and Stanley, 2011; Gaier et al., 2019): we know that the image *can* be represented and that there exists a sequence of mutations that leads to it, but an automated process guided by the pixel difference is unable to find it (Figure 1C). In other words, evolving artifacts with life-like complexity—brains, images, robots—is not only a representation problem.

This is the *stepping stone problem* (Stanley and Lehman, 2015) (Figure 1): the intermediate steps are often very different from the final product; therefore, how can evolution identify the best stepping stones? In the evolutionary computation literature, the fitness landscape is called *deceptive*, which means that the optimization process is "trapped" by attractive local optima that deceive it (Goldberg, 1989). In evolutionary biology, this was already one of the main question addressed to Darwin (1859); for example, "What use is half a wing?," Mivart was asking (Mivart, 1871). The modern answer is "structural shift with structural continuity'' and exaptation (Gould and Vrba, 1982), which means that the same structure is evolved for a function and then coopted (''exapted'') for new functions. Similar questions arise when looking at technological or scientific breakthroughs (Feyerabend, 1975; Stanley and Lehman, 2015), like the common saying that electricity was not invented while working on improving candles. This kind of branching is common in Picbreeder, in which each user can start from the result of others but select images with its own objective. But how to abstract this principle to evolve the behavior of machines with an automated process?

A radical idea is to ignore the fitness function and instead simply generate novel things (Lehman and Stanley, 2011a). The intuition is that once the evolutionary process has exhausted all the simple things to do, it will have to find more and more complex behaviors that will create potential stepping stones. For example, once a biped robot has explored all the ways of falling, walking one step is the most novel behavior, then walking many steps, etc. (Lehman and Stanley, 2011a). In concrete terms, Novelty Search proposes to replace the fitness function by a novelty score, which is computed by comparing the *behavior* of each individuals to those of all, for computational reasons a subset of, the individuals generated so far (stored in a large archive). Counterintuitively, Novelty Search can sometimes find higher-quality solutions or find them quicker than objective-based search (Lehman and Stanley, 2011a). Novelty search also encourages evolvability more than objective-based search (Doncieux et al., 2020; Lehman and Stanley, 2013) because it selects indirectly for the ability to generate novel behaviors with few changes in the genotypes.

An important contribution of Novelty Search is that it introduces a behavior characterization to compare solutions instead of comparing genotypes or only looking at the fitness value. This idea has been successfully exploited to maintain a population in which individuals all behave differently, which improves objective-based evolutionary search in most situations (Mouret, 2011; Mouret and Doncieux, 2012) and bypasses the difficulties in comparing neural networks to maintain a diverse population. In these methods, individuals are ranked using both their fitness and how different they are from the current population (Mouret and Doncieux, 2012) or from the previously generated behaviors (Mouret, 2011). By taking the general objective into account, these methods combine the benefits of Novelty Search while still steering the population toward the goal. It thus can avoid ''wasting'' computing resources when the behavior space is too large to be fully explored. Introducing the concept of behavior in evolutionary algorithms also triggered novel questions about how to characterize them in the most general and task-agnostic way, because choosing a specific behavior distance will constrain what stepping stones will be considered (Doncieux and Mouret, 2010).

## THE NEW BREED: QUALITY DIVERSITY ALGORITHMS

Novelty Search triggered many debates about how to best include objectives when evolving the behavior of machines. On the one hand, the experiments with Novelty Search demonstrate that objectives can be detrimental to the evolutionary process; on the other hand, we are not interested in every behavior and we have limited computational resources: exploring in all directions is likely to be infeasible in many interesting problems. For instance, scientists might explore in many directions, but they usually have some objective in mind; similarly, the users of Picbreeder all have some internal objective, for example, resemblance to known objects, as they are not purely driven by the desire to create a new picture.

The current successors of Novelty Search combine objective-based optimization and novelty-based exploration by searching for a *collection* of solutions that are as diverse and as high-performing as possible. They are called ''Quality Diversity (QD) algorithms'' or ''illumination algorithms'' (Mouret and Clune, 2015; Pugh et al., 2016; Cully and Demiris, 2017).

MAP-Elites (Multidimensional Archive of Phenotypic Elites) (Mouret and Clune, 2015; Vassiliades et al., 2017) is one of the simplest Quality Diversity algorithms, which makes it easy to implement and to modify (Figure 2). This is why it is currently very successful in the evolutionary computation community (see Cully et al., 2020 for a list of quality diversity papers). The general principles of MAP-Elites are:

- the fitness function returns both a fitness value and an n-dimensional behavioral descriptor (or feature descriptor) that describes how the problem is solved;
- the behavior space is divided into behavioral niches (for example, a two-dimensional grid for two-dimensional descriptors); the set of niches is called the map or the archive;

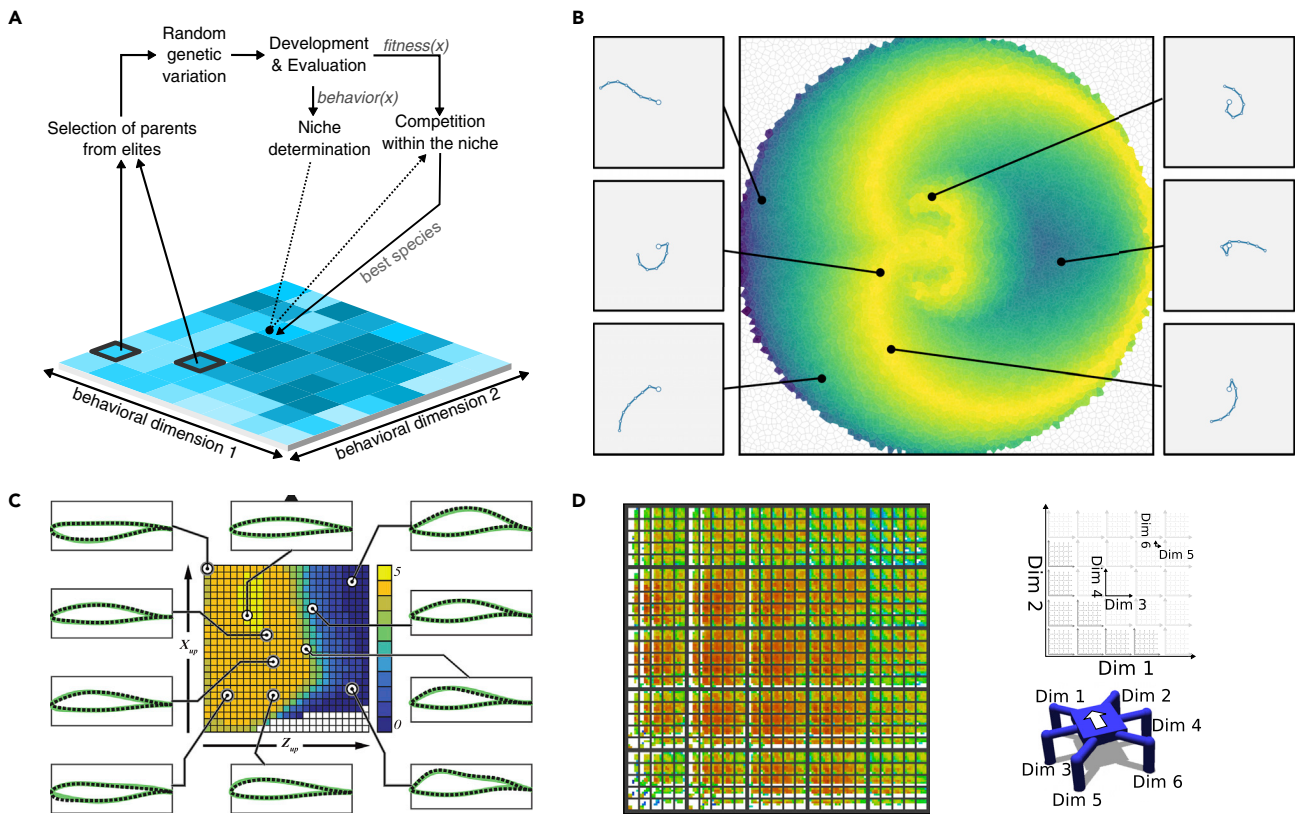

**Figure 2. Quality Diversity Algorithms**
(A) Main loop of MAP-Elites (Mouret and Clune, 2015), one of the most successful quality diversity algorithms.
(B) Example output of MAP-Elites. Each colored cell is a niche in the behavioral space, which was created with a centroidal Voronoi tesselation (Vassiliades et al., 2017). In this experiment, the genotype space is the joint positions of a seven-dimensional planar arm; the behavior space is the position of the end-effector; the fitness is the variance of the joints (to be minimized to favor smooth configurations). The color denotes the fitness (best fitness values are in yellow).
(C) Two-dimensional map of airfoils, adapted from Gaier et al. (2018). We can understand that the feature on the vertical axis ($X_{up}$) has less influence on the fitness than the horizontal one ($Z_{up}$).
(D) Six-dimensional map for six-legged gaits. Each pixel is the best gait for the corresponding amount of time that each foot touched the ground during the evaluation. Adapted from Cully et al. (2015).

- each niche stores the solution with the best known fitness for this particular behavioral descriptors; this solution is called the elite;
- to improve the map, random elites are chosen from the map, variation operators are applied (mutation and crossover), and the new solutions compete with the existing elites according to their behavioral descriptors.

Importantly, the result of a quality diversity algorithm is a collection of high-performing solutions—the map or archive—and not a single ''optimal'' solution.

MAP-Elites can be seen as an abstract model of the evolution of species, that is, macroevolution. In that view, each behavioral niche corresponds to a different ecological niche. In nature, each niche is occupied by a single species, which outperforms all the previous species for this specific niche. Mutants compete inside their niche but not globally with all the other species. For instance, deep sea fish do not compete with tigers: they both are the elite of their niche. However, lions could compete with tigers and could one day replace them after a few adaptations. Similarly, in MAP-Elites, if a genetic variation makes a solution better while behaving similarly, the new solution will replace the current elite (local competition); however, if a variation makes it behave differently, it will compete with the species that currently occupies a different niche and potentially replace it if its fitness is better. To summarize, elites in MAP-Elites can be seen as species that occupy specific ecological niches, so that there is competition inside the niche but no global competition. After all, tigers are not "better" or "fitter" than deep sea fish; they are simply different because they occupy different niches. By contrast, traditional evolutionary algorithms are based on a global competition between all the members of a population so that the process converges to a single, optimum genome, which would correspond to microevolution (intra-species evolution).

In nature, species share a surprisingly high number of genes. For instance, humans and drosophila (fruit flies) share about 60% of their genes (Adams et al., 2000), whereas they occupy very different ecological niches. One way to understand this fact is that human and drosophila actually use similar building blocks but combine them differently, like they both have neurons for information processing. Similarly, high-performing solutions found with quality diversity algorithms are well spread in the behavioral space but often concentrated in a specific "elite hypervolume" in the genotypic space (Vassiliades and Mouret, 2018), which means that their genes are similar. This observation makes the crossover variation operator particularly effective in quality diversity algorithm, in contrast to traditional evolutionary algorithms.

MAP-Elites is only one particular quality diversity algorithm and many variants can be designed with the same goal. Most of them vary along three dimensions (Cully and Demiris, 2017): (1) what kind of archive is used (a grid versus an unstructured archive), (2) how parents are selected (uniformly from the archive, using an auxiliary population, favoring the fittest individuals, etc.), and (3) which variation operators are used. For example, Novelty Search with Local Competition (Lehman and Stanley, 2011b; Cully and Mouret, 2016), which might have been the first quality diversity algorithm, stores solutions in an unstructured archive by adding them depending on their behavioral distance with the current solutions of the archive and selects parents from an auxiliary population sorted using both fitness and novelty with regards to the archive. Many recent contributions improve current algorithms (especially MAP-Elites) by making them faster (fewer generations for the same quality) and more scalable (both for higher-dimensional search space and higher-dimensional behavior space). For instance, they proposed advanced variation operators (Vassiliades and Mouret, 2018; Fontaine et al., 2020; Colas et al., 2020) and scalable definitions of behavioral niches (Vassiliades et al., 2017). Promising progress has also been made to handle noisy fitness functions and noisy descriptors (Flageat and Cully, 2020) and to find interesting behavioral dimensions automatically (Cully, 2019).

Quality diversity algorithms like MAP-Elites are therefore a fresh and thought-stimulating view at artificial evolution that takes inspiration from macroevolution (evolution of species) instead of microevolution (intraniche evolution). This makes them better at generating interesting stepping stones and at imitating the astonishing creativity of natural evolution. However, they are not mathematical optimization algorithms anymore and the community is still exploring the many situations for which it is useful to generate thousands of high-quality solutions.

One of the most prominent illustrations of MAP-Elites is to generate gait repertoires for legged robots (Cully et al., 2015; Pautrat et al., 2018) (Figure 2D). Cully et al. (2015) thus leveraged MAP-Elites to create about 15,000 different six-legged gaits that differ by how much each foot is in contact with the floor. In that case, the behavioral descriptor is six-dimensional, percentage of contact for each of the six legs, the gaits are described by 36 parameters (3 parameters for each of the 12 controlled joints), and the fitness function is the distance covered during the simulation of the gait for 5 s. After a few million evaluations in simulation, MAP-Elites finds high-performing gaits that rely differently on each leg, for instance, the best gait when using the six legs, the best gait when not using much the front leg, and so forth. Similar results were obtained with different behavior descriptions, like the orientation of the body. Since the gaits are evolved in simulation, the algorithm can run for a long time (a few days) on a large parallel computer to get high-quality results.

If the real robot is damaged (for example, a leg is broken), it draws from this precomputed repertoire an alternative gait that works in spite of the damage by searching by trial and error with an algorithm called Bayesian optimization (Shahriari et al., 2015). This approach allows a six-legged robot to recover from many damage conditions in less than 2 min (a dozen of trials of 5 s). An interesting parallel can be drawn with extinction events in biology (like the extinction of dinosaurs): by evolving diverse species of solutions with a quality diversity algorithm, it is likely that some of them will still survive after a large change in the environment (here, a damage to the robot). Recent extensions generate gaits to reach each point around the robot (Duarte et al., 2017; Chatzilygeroudis et al., 2018; Kaushik et al., 2020); in that case, the behavior description is the position of the robot in the plane after 5 s (two-dimensional) and these elementary behaviors can then be chained with a planning algorithm to achieve complex trajectories (Chatzilygeroudis et al., 2018; Kaushik et al., 2020).

Quality diversity algorithms were initially introduced in the robotics and artificial intelligence community. However, many other fields benefit from algorithms that can propose a collection of interesting solutions to a problem and let users pick up the one they like; using their own criteria like aesthetics, manufacturing difficulty or gameplay. Most interestingly, this approach avoids the need to encode all the requirements in the fitness function, which is required when using traditional optimization algorithms. So far, promising results have been published for designing three-dimensional shapes of aerodynamic bikes (Gaier et al., 2018) (Figure 2C), to generate content for different games (Fontaine et al., 2020; Gravina et al., 2019), to design molecules (Verhellen and Van den Abeele, 2020), to solve workforce scheduling and routing problems (Urquhart and Hart, 2018), and to find adversarial examples for deep neural networks (Nguyen et al., 2015) or malware (Babaagba et al., 2020).

## CONCLUSION

Evolution continuously innovates by creating new species in unexpected ways (Blount et al., 2008), but such creativity is rarely observed when using traditional evolutionary algorithms. One main reason is that these algorithms are close to microevolution, that is, intra-niche evolution, because they aim at finding a single, optimal genome. Quality diversity algorithms are closer to macroevolution by explicitly implementing behavioral niches. More generally, quality diversity algorithms search for many ways of maximizing a function, in the same way as natural evolution searches for many ways of maximizing the reproductive success, the fitness, one for each specific ecological environment.

Quality diversity algorithms are, however, a rough abstraction of macroevolution. In particular, they currently require the crafting of a function that decides to which ecological niche each behavior belongs. This function cannot measure what it was not designed for and therefore can miss truly novel innovations. For instance, if it looks at the two-dimensional trajectories of robots, then it will never be able to identify that a flying robot behaves very differently, as it only ''sees'' the projection of the trajectory to the ground. In that situation, the experimenter has to anticipate that robots could fly and compare three-dimensional trajectories: this severely limits the creativity of the process. In other words, quality diversity algorithms are not fully *open-ended* (Stanley, 2019). Instead, they are balancing open-endedness with practical use (and implementation) by being ''more open-ended'' than traditional evolutionary algorithms. An important research avenue is how to design such algorithms so that they are more open-ended while still have practical uses.

This article looks at machines only through the lens of their behavior. Nevertheless, a given morphology always constrains the possible behaviors and therefore limits the creativity of the process. In addition, it is often possible to simplify the control policy by using a different morphology (Pfeifer and Bongard, 2006). Overall, most of the ideas described here can be extended to ''co-evolve'' both the morphology and the controller, but the difficulty is increased because the search space is much bigger. To break the complexity, a promising approach is to leverage diversity algorithms to create ''catalogs'' of parts for each level of a robot, from materials to control (Howard et al., 2019): a catalog of material with diverse features, then a catalog of parts that combine these materials, and so forth.

## ACKNOWLEDGMENTS

This work received funding from the European Research Council (ERC) under the European Union's Horizon 2020 research and innovation program (GA no. 637972, project ''ResiBots'') and the Lifelong Learning Machines program (L2M) from DARPA/MTO under Contract No. FA8750-18-C-0103. The author thanks Adam Gaier for his help with preparing this manuscript.

## AUTHOR CONTRIBUTIONS

J.-B.M. wrote the paper.

## REFERENCES

Aaltonen, T., Adelman, J., Akimoto, T., Albrow, M., González, B.Á., Amerio, S., Amidei, D., Anastassov, A., Annovi, A., Antos, J., et al. (2009). Measurement of the top-quark mass with dilepton events selected using neuroevolution at cdf. Phys. Rev. Lett. *102*, 152001.

Adams, M.D., Celniker, S.E., Holt, R.A., Evans, C.A., Gocayne, J.D., Amanatides, P.G., Scherer, S.E., Li, P.W., Hoskins, R.A., Galle, R.F., et al. (2000). The genome sequence of drosophila melanogaster. Science *287*, 2185–2195.

Ahnert, S.E. (2017). Structural properties of genotype–phenotype maps. J. R. Soc. Interfaces *14*, 20170275.

K.O. Babaagba, Z. Tan, and E. Hart. Automatic generation of adversarial metamorphic malware using map-elites. In International Conference on the Applications of Evolutionary Computation (Part of EvoStar), pages 117–132. Springer, 2020.

Banzhaf, W., Nordin, P., Keller, R.E., and Francone, F.D. (1998). Genetic Programming (Springer).

Blount, Z.D., Borland, C.Z., and Lenski, R.E. (2008). Historical contingency and the evolution of a key innovation in an experimental population of *Escherichia coli*. Proc. Natl. Acad. Sci. U S A *105*, 7899–7906.

Bongard, J., Zykov, V., and Lipson, H. (2006). Resilient machines through continuous self-modeling. Science *314*, 1118–1121.

Bongard, J.C., and Pfeifer, R. (2003). Evolving complete agents using artificial ontogeny. In Morpho-functional Machines: The New Species, F. Hara and R. Pfeifer, eds. (Springer), pp. 237–258.

Braitenberg, V. (1986). Vehicles: Experiments in Synthetic Psychology (MIT press).

Brodbeck, L., Hauser, S., and Iida, F. (2015). Morphological evolution of physical robots through model-free phenotype development. PLoS One *10*, e0128444.

Cangelosi, A., Parisi, D., and Nolfi, S. (1994). Cell division and migration in a 'genotype' for neural networks. Netw. Comput. Neural Syst. *5*, 497–515.

Chatzilygeroudis, K., Vassiliades, V., and Mouret, J.-B. (2018). Reset-free trial-and-error learning for robot damage recovery. Robot. Auton. Syst. *100*, 236–250.

Clune, J., Stanley, K.O., Pennock, R.T., and Ofria, C. (2011). On the performance of indirect encoding across the continuum of regularity. IEEE Trans. Evol. Comput. *15*, 346–367.

Clune, J., Mouret, J.-B., and Lipson, H. (2013). The evolutionary origins of modularity. Proc. R. Soc. B Biol. Sci. *280*, 20122863.

C. Colas, J. Huizinga, V. Madhavan, and J. Clune. Scaling map-elites to deep neuroevolution. In Proceedings of the 2020 Genetic and Evolutionary Computation Conference, 2020.

Cully, A. (2019). Autonomous skill discovery with quality-diversity and unsupervised descriptors. In Proceedings of the Genetic and Evolutionary Computation Conference, pp. 81–89.

Cully, A., and Demiris, Y. (2017). Quality and diversity optimization: a unifying modular framework. IEEE Trans. Evol. Comput. *22*, 245–259.

Cully, A., and Mouret, J.-B. (2016). Evolving a behavioral repertoire for a walking robot. Evol. Comput. *24*, 59–88.

Cully, A., Clune, J., Tarapore, D., and Mouret, J.-B. (2015). Robots that can adapt like animals. Nature *521*, 503–507.

Cully, A., et al. (2020). Quality-diversity optimisation algorithms. https://quality-diversity.github.io.

Cussat-Blanc, S., Harrington, K., and Banzhaf, W. (2019). Artificial gene regulatory networks—a review. Artif. Life *24*, 296–328.

Darwin, C. (1859). On the Origin of Species (John Murray).

Dasgupta, D., and McGregor, D.R. (1992). Designing application-specific neural networks using the structured genetic algorithm. In [Proceedings] COGANN-92: International Workshop on Combinations of Genetic Algorithms and Neural Networks (IEEE), pp. 87–96.

Davidson, E.H. (2010). The Regulatory Genome: Gene Regulatory Networks in Development and Evolution (Elsevier).

De Jong, K.A. (2016). Evolutionary Computation: A Unified Approach (MIT Press).

Deb, K., and Beyer, H.-G. (2001). Self-adaptive genetic algorithms with simulated binary crossover. Evol. Comput. *9*, 197–221.

Doncieux, S., and Mouret, J.-B. (2010). Behavioral diversity measures for evolutionary robotics. In IEEE Congress on Evolutionary Computation (IEEE), pp. 1–8.

Doncieux, S., and Mouret, J.-B. (2014). Beyond black-box optimization: a review of selective pressures for evolutionary robotics. Evol. Intell. *7*, 71–93.

S. Doncieux, G. Paolo, A. Laflaquière, and A. Coninx. Novelty search makes evolvability inevitable. In Proceedings of the Genetic and Evolutionary Computation Conference (GECCO), 2020.

Duarte, M., Gomes, J., Oliveira, S.M., and Christensen, A.L. (2017). Evolution of repertoire-based control for robots with complex locomotor systems. IEEE Trans. Evol. Comput. *22*, 314–328.

Eggenberger, P. (1997). Evolving morphologies of simulated 3d organisms based on differential gene expression. In Fourth European Conference on Artificial Life, 4Fourth European Conference on Artificial Life (MIT Press), p. 205.

Erwin, D.H. (2000). Macroevolution is more than repeated rounds of microevolution. Evol. Dev. *2*, 78–84.

Feyerabend, P. (1975). Against Method: Outline of an Anarchistic Theory of Knowledge (Verso).

M. Flageat and A. Cully. Fast and stable map-elites in noisy domains using deep grids. In Proceedings of the 2020 Conference on Artificial Life, 2020.

Floreano, D., and Mondada, F. (1998). Evolutionary neurocontrollers for autonomous mobile robots. Neural Netw. *11*, 1461–1478.

Floreano, D., Dürr, P., and Mattiussi, C. (2008). Neuroevolution: from architectures to learning. Evol. Intell. *1*, 47–62.

Fontaine, M.C., Togelius, J., Nikolaidis, S., and Hoover, A.K. (2020). Covariance matrix adaptation for the rapid illumination of behavior space. In Proceedings of the 2020 Genetic and Evolutionary Computation Conference, pp. 94–102.

Gaier, A., Asteroth, A., and Mouret, J.-B. (2018). Data-efficient design exploration through surrogate-assisted illumination. Evolutionary computation *26*, 381–410.

A. Gaier, A. Asteroth, and J.-B. Mouret. Are quality diversity algorithms better at generating stepping stones than objective-based search? In Proceedings of the Genetic and Evolutionary Computation Conference Companion, pages 115–116, 2019.

Gauci, J., and Stanley, K.O. (2008). A case study on the critical role of geometric regularity in machine learning. In AAAI, pp. 628–633.

Goldberg, D.E. (1989). Genetic algorithms and Walsh functions: Part II, deception and its analysis. Complex Syst. *3*, 153–171.

Gould, S.J., and Vrba, E.S. (1982). Exaptation – a missing term in the science of form. Paleobiology *8*, 4–15.

D. Gravina, A. Khalifa, A. Liapis, J. Togelius, and G.N. Yannakakis. Procedural content generation through quality diversity. In IEEE Conference on Games (CoG), 1–8. IEEE, 2019.

Gruau, F. (1994). Automatic definition of modular neural networks. Adapt. Behav. *3*, 151–183.

Hansen, N., Müller, S.D., and Koumoutsakos, P. (2003). Reducing the time complexity of the derandomized evolution strategy with covariance matrix adaptation (CMA-ES). Evol. Comput. *11*, 1–18.

Hausknecht, M., Lehman, J., Miikkulainen, R., and Stone, P. (2014). A neuroevolution approach to general atari game playing. IEEE Trans. Comput. Intell. AI Games *6*, 355–366.

Heess, N., TB, D., Sriram, S., Lemmon, J., Merel, J., Wayne, G., Tassa, Y., Erez, T., Wang, Z., Eslami, S., et al. (2017). Emergence of locomotion behaviours in rich environments. arXiv. arXiv:1707.02286.

Heidrich-Meisner, V., and Igel, C. (2009). Neuroevolution strategies for episodic reinforcement learning. J. Algorithms *64*, 152–168.

Hornby, G.S., and Pollack, J.B. (2001). Evolving l-systems to generate virtual creatures. Comput. Graph. *25*, 1041–1048.

Hornby, G.S., Takamura, S., Yamamoto, T., and Fujita, M. (2005). Autonomous evolution of dynamic gaits with two quadruped robots. IEEE Trans. Robot. *21*, 402–410.

Howard, D., Eiben, A.E., Kennedy, D.F., Mouret, J.-B., Valencia, P., and Winkler, D. (2019). Evolving embodied intelligence from materials to machines. Nat. Mach. Intell. *1*, 12–19.

J. Huizinga, J. Clune, and J.-B. Mouret. Evolving neural networks that are both modular and regular: hyperneat plus the connection cost technique. In Proceedings of the 2014 Annual Conference on Genetic and Evolutionary Computation, pages 697–704, 2014.

Jakobi, N. (1997). Evolutionary robotics and the radical envelope-of-noise hypothesis. Adapt. Behav. *6*, 325–368.

Jelisavcic, M., De Carlo, M., Hupkes, E., Eustratiadis, P., Orlowski, J., Haasdijk, E., Auerbach, J.E., and Eiben, A.E. (2017). Real-world evolution of robot morphologies: a proof of concept. Artif. Life *23*, 206–235.

Kashtan, N., and Alon, U. (2005). Spontaneous evolution of modularity and network motifs. Proc. Natl. Acad. Sci. U S A *102*, 13773–13778.

Kaushik, R., Desreumaux, P., and Mouret, J.-B. (2020). Adaptive prior selection for repertoire-based online adaptation in robotics. Front. Robot. AI *6*, 151.

Kitano, H. (1990). Designing neural networks using genetic algorithms with graph generation system. Complex Syst. *4*, 461–476.

Kochenderfer, M.J., and Wheeler, T.A. (2019). Algorithms for Optimization (MIT Press).

Kodjabachian, J., and Meyer, J.-A. (1998). Evolution and development of neural controllers for locomotion, gradient-following, and obstacle-avoidance in artificial insects. IEEE Trans. Neural Netw. *9*, 796–812.

Koos, S., Mouret, J.-B., and Doncieux, S. (2012). The transferability approach: crossing the reality gap in evolutionary robotics. IEEE Trans. Evol. Comput. *17*, 122–145.

Koza, J.R. (1992). Genetic Programming: On the Programming of Computers by Means of Natural Selection, *1* (MIT press).

Kriegman, S., Blackiston, D., Levin, M., and Bongard, J. (2020). A scalable pipeline for designing reconfigurable organisms. Proc. Natl. Acad. Sci. U S A *117*, 1853–1859.

Lehman, J., and Stanley, K.O. (2011a). Abandoning objectives: evolution through the search for novelty alone. Evol. Comput. *19*, 189–223.

J. Lehman and K.O. Stanley. Evolving a diversity of virtual creatures through novelty search and local competition. In Proceedings of the 13th Annual Conference on Genetic and Evolutionary Computation, pages 211–218, 2011b.

Lehman, J., and Stanley, K.O. (2013). Evolvability is inevitable: Increasing evolvability without the pressure to adapt. PLoS One *8*, e62186.

Lehman, J., Clune, J., Misevic, D., Adami, C., Altenberg, L., Beaulieu, J., Bentley, P.J., Bernard, S., Beslon, G., Bryson, D.M., et al. (2020). The surprising creativity of digital evolution: a collection of anecdotes from the evolutionary computation and artificial life research communities. Artif. Life *26*, 274–306.

Lindenmayer, A. (1968). Mathematical models for cellular interactions in development ii. simple and branching filaments with two-sided inputs. J. Theor. Biol. *18*, 300–315.

Lipson, H. (2007). Principles of modularity, regularity, and hierarchy for scalable systems. J. Biol. Phys. Chem. *7*, 125.

Lipson, H., and Pollack, J.B. (2000). Automatic design and manufacture of robotic lifeforms. Nature *406*, 974–978.

Mania, H., Guy, A., and Recht, B. (2018). Simple random search of static linear policies is competitive for reinforcement learning. In Advances in Neural Information Processing Systems, pp. 1800–1809.

J.F. Miller. Evolving developmental programs for adaptation, morphogenesis, and self-repair. In European Conference on Artificial Life, pages 256–265. Springer, 2003.

Mivart, S.G.J. (1871). On the Genesis of Species (Appleton).

Mnih, V., Kavukcuoglu, K., Silver, D., Rusu, A.A., Veness, J., Bellemare, M.G., Graves, A., Riedmiller, M., Fidjeland, A.K., Ostrovski, G., et al. (2015). Human-level control through deep reinforcement learning. Nature *518*, 529–533.

Mouret, J.-B. (2011). Novelty-based multiobjectivization. In New Horizons in Evolutionary Robotics (Springer), pp. 139–154.

Mouret, J.-B., and Clune, J. (2015). Illuminating search spaces by mapping elites. arXiv. arXiv:1504.04909.

Mouret, J.-B., and Doncieux, S. (2008). MENNAG: a modular, regular and hierarchical encoding for neural-networks based on attribute grammars. Evol. Intell. *1*, 187–207.

Mouret, J.-B., and Doncieux, S. (2012). Encouraging behavioral diversity in evolutionary robotics: an empirical study. Evol. Comput. *20*, 91–133.

Mouret, J.-B., and Tonelli, P. (2014). Artificial evolution of plastic neural networks: a few key concepts. In Growing Adaptive Machines, T. Kowaliw, N. Bredeche, and R. Doursat, eds. (Springer), pp. 251–261.

A. Nguyen, J. Yosinski, and J. Clune. Deep neural networks are easily fooled: high confidence predictions for unrecognizable images. In Proceedings of the IEEE Conference on Computer Vision and Pattern Recognition, pages 427–436, 2015.

Nolfi, S., Floreano, D., and Floreano, D.D. (2000). Evolutionary Robotics: The Biology, Intelligence, and Technology of Self-Organizing Machines (MIT press).

R. Pautrat, K. Chatzilygeroudis, and J.-B. Mouret. Bayesian optimization with automatic prior selection for data-efficient direct policy search. In 2018 IEEE International Conference on Robotics and Automation (ICRA), pages 7571–7578. IEEE, 2018.

Penco, L., Hoffman, E.M., Modugno, V., Gomes, W., Mouret, J.-B., and Ivaldi, S. (2020). Learning robust task priorities and gains for control of redundant robots. IEEE Robot. Autom. Lett. *5*, 2626–2633.

Pfeifer, R., and Bongard, J. (2006). How the Body Shapes the Way We Think: A New View of Intelligence (MIT press).

Pigliucci, M. (2008). Is evolvability evolvable? Nat. Rev. Genet. *9*, 75–82.

Pugh, J.K., Soros, L.B., and Stanley, K.O. (2016). Quality diversity: a new frontier for evolutionary computation. Front. Robot. AI *3*, 40.

Reznick, D.N., and Ricklefs, R.E. (2009). Darwin's bridge between microevolution and macroevolution. Nature *457*, 837–842.

Salimans, T., Ho, J., Chen, X., Sidor, S., and Sutskever, I. (2017). Evolution strategies as a scalable alternative to reinforcement learning. arXiv. arXiv:1703.03864.

Secretan, J., Beato, N., D'Ambrosio, D.B., Rodriguez, A., Campbell, A., Folsom-Kovarik, J.T., and Picbreeder, K.O. Stanley. (2011). A case study in collaborative evolutionary exploration of design space. Evol. Comput. *19*, 373–403.

Shahriari, B., Swersky, K., Wang, Z., Adams, R.P., and De Freitas, N. (2015). Taking the human out of the loop: a review of Bayesian optimization. Proc. IEEE *104*, 148–175.

K. Sims. Artificial evolution for computer graphics. In Proceedings of the 18th Annual Conference on Computer Graphics and Interactive Techniques, pages 319–328, 1991.

K. Sims. Evolving virtual creatures. In SIGGRAPH '94: Proceedings of the 21st Annual Conference on Computer Graphics and Interactive Techniques, pages 15–22, 1994.

Soltoggio, A., Stanley, K.O., and Risi, S. (2018). Born to learn: the inspiration, progress, and future of evolved plastic artificial neural networks. Neural Netw. *108*, 48–67.

Southan, C. (2004). Has the yo-yo stopped? an assessment of human protein-coding gene number. Proteomics *4*, 1712–1726.

Stanley, K.O. (2007). Compositional pattern producing networks: a novel abstraction of development. Genet. Program. Evol. Mach. *8*, 131–162.

Stanley, K.O. (2019). Why open-endedness matters. Artif. Life *25*, 232–235.

Stanley, K.O., and Lehman, J. (2015). Why Greatness Cannot Be Planned: The Myth of the Objective (Springer).

Stanley, K.O., and Miikkulainen, R. (2002). Evolving neural networks through augmenting topologies. Evol. Comput. *10*, 99–127.

Stanley, K.O., and Miikkulainen, R. (2003). A taxonomy for artificial embryogeny. Artif. Life *9*, 93–130.

Stanley, K.O., Bryant, B.D., and Miikkulainen, R. (2005). Real-time neuroevolution in the nero video game. IEEE Trans. Evol. Comput. *9*, 653–668.

Stanley, K.O., Ambrosio, D.B.D., and Gauci, J. (2009). A hypercube-based encoding for evolving large-scale neural networks. Artif. Life *15*, 185–212.

Stanley, K.O., Clune, J., Lehman, J., and Miikkulainen, R. (2019). Designing neural networks through neuroevolution. Nat. Machine Intelligence *1*, 24–35.

Such, F.P., Madhavan, V., Conti, E., Lehman, J., Stanley, K.O., and Clune, J. (2017). Deep neuroevolution: genetic algorithms are a competitive alternative for training deep neural networks for reinforcement learning. arXiv. arXiv:1712.06567.

Sutton, R.S., and Barto, A.G. (2018). Reinforcement Learning: An Introduction (MIT press).

Tan, J., Zhang, T., Coumans, E., Iscen, A., Bai, Y., Hafner, D., Bohez, S., and Vanhoucke, V. (2018). Sim-to-real: learning agile locomotion for quadruped robots. In Proceedings of Robotics: Science and Systems, Pittsburgh, Pennsylvania. https://doi.org/10.15607/RSS.2018.XIV.010.

J. Tobin, R. Fong, A. Ray, J. Schneider, W. Zaremba, and P. Abbeel. Domain randomization for transferring deep neural networks from simulation to the real world. In 2017 IEEE/RSJ International Conference on Intelligent Robots and Systems (IROS), pages 23–30. IEEE, 2017.

Togelius, J., Schaul, T., Wierstra, D., Igel, C., Gomez, F., and Schmidhuber, J. (2009). Ontogenetic and phylogenetic reinforcement learning. Künstliche Intelligenz *23*, 30–33.

Turing, A.M. (1952). The chemical basis of morphogenesis. Phil. Trans. R. Soc. Lond. Ser. B Biol. Sci. *237*, 37–72.

N. Urquhart and E. Hart. Optimisation and illumination of a real-world workforce scheduling and routing application (wsrp) via map-elites. In International Conference on Parallel Problem Solving from Nature, pages 488–499. Springer, 2018.

Vassiliades, V., and Mouret, J.-B. (2018). Discovering the elite hypervolume by leveraging interspecies correlation. In Proceedings of the Genetic and Evolutionary Computation Conference, pp. 149–156.

Vassiliades, V., Chatzilygeroudis, K., and Mouret, J.-B. (2017). Using centroidal voronoi tessellations to scale up the Multidimensional Archive of Phenotypic elites algorithm. IEEE Trans. Evol. Comput. *22*, 623–630.

Verhellen, J., and Van den Abeele, J. (2020). Illuminating elite patches of chemical space. Chem. Sci. *4*, 120–131.

Wagner, G.P., Pavlicev, M., and Cheverud, J.M. (2007). The road to modularity. Nat. Rev. Genet. *8*, 921–931.

Whiteson, S. (2012). Evolutionary computation for reinforcement learning. In Reinforcement Learning, M. Wiering and M.V. Otterlo, eds. (Springer), pp. 325–355.

B.G. Woolley and K.O. Stanley. On the deleterious effects of a priori objectives on evolution and representation. In Proceedings of the 13th Annual Conference on Genetic and Evolutionary Computation, pages 957–964, 2011.